# SolarSAM: Building-scale Photovoltaic Potential Assessment Based on Segment Anything Model (SAM) and Remote Sensing for Emerging City

**Guohao Wang**

Department of Power and Electrical Engineering, Northwest A&F University, Yangling 712100, China;

E-mail address: guohaowang@ieee.org (G. Wang)

**Abstract:** Driven by advancements in photovoltaic (PV) technology, solar energy has emerged as a promising renewable energy source, due to its ease of integration onto building rooftops, facades, and windows. For the emerging cities, the lack of detailed street-level data presents a challenge for effectively assessing the potential of building-integrated photovoltaic (BIPV). To address this, this study introduces SolarSAM, a novel BIPV evaluation method that leverages remote sensing imagery and deep learning techniques, and an emerging city in northern China is utilized to validate the model performance. During the process, SolarSAM segmented various building rooftops using text prompt guided semantic segmentation. Separate PV models were then developed for Rooftop PV, Facade-integrated PV, and PV windows systems, using this segmented data and local climate information. The potential for BIPV installation, solar power generation, and city-wide power self-sufficiency were assessed, revealing that the annual BIPV power generation potential surpassed the city's total electricity consumption by a factor of 2.5. Economic and environmental analysis were also conducted, including levelized cost of electricity and carbon reduction calculations, comparing different BIPV systems across various building categories. These findings demonstrated the model's performance and reveled the potential of BIPV power generation in the future.

**Highlights:**

1. A BIPV potential assessment method with SAM and satellite imagery is first presented.

2. Economic analysis of rooftop, facade, and window integrated-PV on various buildings was taken.

3. The annual potential power generation from BIPV in Zibo is 2.5 times the annual power consumption.

4. Environmental analysis reveals the annual potential carbon reduction is $7.08 \times 10^7$ T $CO_2$.

**Nomenclature:**

| Symbol | Description |
|---|---|
| $P_i$ | Set of pixels belonging to the mask of building category $i$ |
| $P_b$ | Set of pixels in the mask of all the buildings |
| $S$ | Scale factor converting pixels to real-world area units |
| $A_i$ | Area of the building category $i$ |
| $AAPV_i$ | Actual area available for PV installation from building category $i$ |
| $A/RA$ | Relation of area between the area of wall or windows and rooftop |
| $K_{mapping}$ | Correction mapping factor |
| $E_t$ | Annual PV electricity generation at t year (kWh) |
| $C_t$ | Annual total cost (CNY) |
| $C_{start}$ | Initial investment for purchasing and installing PV hardware |
| $C_{O\&M}$ | Fee for the operating and maintenance of the PV system |
| $C_{rent}$ | Cost of renting rooftops, walls, and windows |
| $r$ | Discount rate |
| $t$ | Year |
| $n$ | Number of years |
| $Capacity$ | Installed PV capacity (W) |

| $E_{pv}$ | Electricity generated by BIPV systems (kWh) |
|---|---|
| $EF$ | Carbon emission factor |
| $TP$ | True Positives: pixels correctly classified as belonging to this building category |
| $TN$ | True Negatives: pixels correctly classified as not belonging to this building category |
| $FP$ | False Positives: pixels incorrectly classified as belonging to this building category |
| $FN$ | False Negatives: pixels incorrectly classified as not belonging to this building category |
| $PA$ | the proportion of pixels that are correctly classified by the model |
| $IoU$ | the proportion of the intersection area between the predicted segmentation and the ground truth segmentation to the union area |

# 1 Introduction

The burgeoning global energy demand, coupled with the detrimental environmental impact of fossil fuel reliance, necessitates a paradigm shift towards sustainable energy sources [1]. PV technology has emerged as a promising solution, offering clean and renewable energy with decreasing costs [2,3]. However, traditional ground-mounted PV installations often face challenges related to land scarcity, transmission losses, and integration into existing infrastructure [4]. Building-integrated photovoltaic (BIPV) presents a compelling alternative, seamlessly integrating PV systems into the fabric of urban buildings [5], thus maximizing land use efficiency [6], offering a sustainable and clean energy solution [7], and enabling localized energy generation [8]. Evaluating the potential of BIPV systems in urban environments is crucial for charting a path toward sustainable urban energy landscapes.

BIPV can be integrated into building rooftops [9], facades [10], and windows [11], and the formed Rooftop PV, Facade-integrated PV, and PV window systems present diverse possibilities [12,13]. Rooftop PV systems, utilizing the vast expanse of rooftops, represent the most established and widely adopted BIPV technology [14]. Facade-integrated PV systems transform building facades into energy generators, offering additional surface area for PV installations, particularly in densely built environments [15]. Meanwhile, PV windows, though relatively nascent, offer an innovative approach to harnessing solar energy while preserving building aesthetics and functionality [16]. Consequently, the distinct characteristics of these three BIPV systems, when applied to various building types, result in diverse patterns of energy generation, consumption, and cost-effectiveness. Therefore, assessing the potential of these diverse BIPVs across various building types is crucial for a comprehensive evaluation of urban PV potential.

In the decades, with development of Artificial intelligence (AI), machine learning (ML) and deep learning (DL) has shown their talent in PV potential assessment [19], fault detection [17,18], and power generation forecasting [20,21]. Several studies have explored AI methods for Rooftop PV potential assessment in urban environments [22]. Among them, methods based on geographic information systems (GIS) [23] are the most widely utilization. For instance, Gagnon et al. [24] applied a Multiple Linear Regression (MLR) model with GIS data on building characteristics and meteorological datasets to estimate rooftop PV potential at the US national level. Walch et al. [25,26] investigated the use of RF and Extremely Randomized Trees (ELME) algorithms with LiDAR data and meteorological datasets to classify rooftop shapes and estimate Rooftop PV potential in Switzerland. For the potential assessment of different types of Facade-integrated PV and PV windows, models utilizing both real-world data and simulation methods are proposed. Sun et al. present mapping tool incorporating visibility analysis and solar energy potential assessment to investigate the feasibility of BIPV deployment in densely built-up urban environments [6]. Hadi et al. proposed a novel dynamic simulation model to assess the impact of Facade-integrated PV on electrical production [27], Vulkan et al. assessed the electricity generation potential of BIPV in dense urban environments, utilizing 3-dimensional (3D) modeling to assess the solar potential of various residential building typologies and facade orientations [28]. However, these models still face limitations, such as evaluating only single types of BIPV and requiring strict data conditions. Therefore, a model with simple data requirements and capable of the potential assessment of multiple BIPVs is worth exploring.

In response to the limitations of traditional GIS-based methods for BIPV potential assessment in rapidly growing urban areas, researchers have turned to remote sensing image analysis as a valuable source of data for large-scale solar resource evaluation. For example, TransPV [29] is a computer vision model with state-of-the-art (SOTA) performance on Heilbronn datasets, and GenPV [30] is a neural network-based model for accurate PV panel segmentation on imbalanced datasets. However, as these models focus solely on segmenting and recognizing existing PV panels, they cannot assess the solar potential of buildings without PV panels installed. This presents a challenge for emerging cities where BIPV has yet to see widespread adoption. Therefore, models capable of segmenting and recognizing building structures, rather than just existing panels, hold significant value for assessing BIPV potential. Some building data-driven approaches can assess building photovoltaic potential. For example, Eicker et al. present a model to assess solar energy resources using 3D city models [31]. To assess rooftop and facade solar photovoltaic potential in rural areas, Liu et al. developed a GIS-based approach utilizing 3D building models [32]. However, these methods rely on 3D building data for the whole city, which is often unavailable for emerging cities. Therefore, researchers provide computer vision (CV) models with satellite imagery input and assess the BIPV potential with building segmentation results [33,34].

However, these models can only segment building rooftops and cannot identify building types. Due to variations in building height, different building types have significantly different installation potential for Facade-integrated PV and PV windows [35]. Therefore, these methods are not accurate in assessing the potential of various BIPV applications in emerging cities. Recently, with the development of CV, Large models have shown their talent in zero-short prediction [36,37], and these models can segment the object in one figure with a specific text prompt [38]. Among all the models, SAM [39] which was presented in 2023 has been proven to be capable for the segment of medical images [40], agriculture images [41], and mechanical images [42], this provides an approach for more accurate PV potential assessment.

To address these limitations, this study proposes a novel city-scale BIPV evaluation method leveraging the SAM. Utilizing this method, this paper analyzes the photovoltaic potential, economic feasibility, and environmental impact of an emerging city. The workflow is illustrated in Figure 1. First, satellite imagery and relevant data for an emerging city are collected and organized. The satellite imagery is divided into uniformly sized images, and the Zibo24 dataset is formatted. From this dataset, 100 images are given segmentation labels by humans, and they are selected for hyperparameter tuning and prompt text determination for the SAM. Second, using the optimized model parameters, SAM segments all images, and the rooftop areas of different building types are calculated based on the image scale. Third, the effective installation area for PV systems on various building types is calculated using conversion factors. Finally, a comprehensive analysis is conducted, evaluating city-wide PV installation, electricity generation, economic feasibility, and environmental impact in terms of carbon emission reduction. The contributions of this paper are as follows:

- This paper proposes a deep learning approach for evaluating the potential of integrating PV systems onto rooftops, facades, and windows in emerging cities. This method provides a comprehensive assessment of the overall BIPV capacity.

- This study identifies optimal text prompts and hyperparameter settings for segmenting roofs of different building types in satellite imagery using the SAM model. Segmentation evaluation metrics demonstrate the model's effectiveness in segmenting buildings. The associated code has been open-sourced.

- This paper evaluates and compares the installation potential of various BIPVs to different building categories in an emerging city. Economic feasibility and carbon emission reduction analyses were conducted based on the estimated power generation potential from PV models. The findings provide policy recommendations to promote BIPV installations in this city.

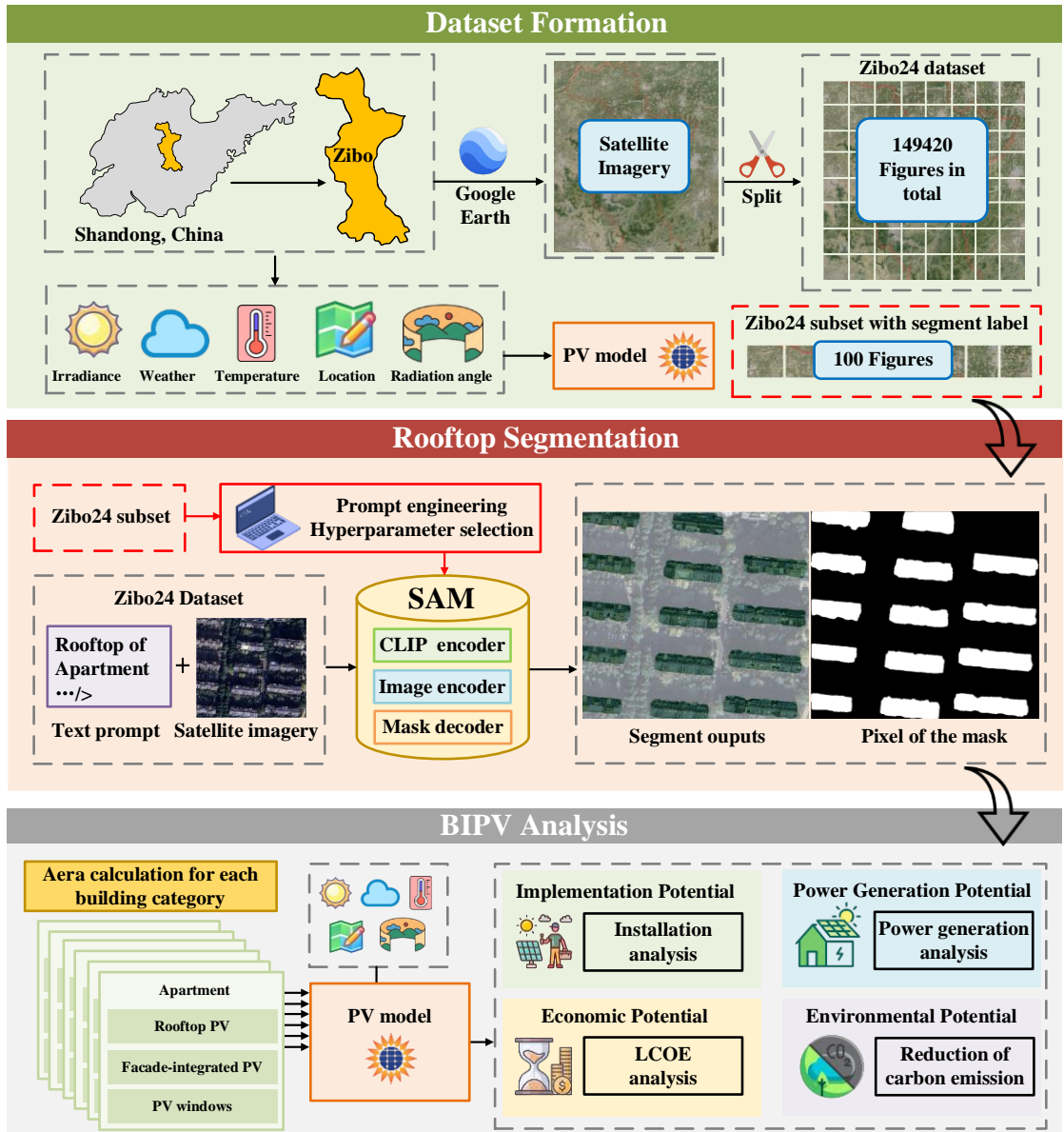

**Figure 1. Workflow of this paper and SolarSAM.**

# 2 Data Acquisition

## 2.1 Six building categories

Zibo, an emerging city in the center of Shandong, China, has witnessed a recent trend of transitioning from an industrial city towards a tourism-based economy. This shift has resulted in a diverse range of building types within the city. For PV potential assessment, buildings in this city can be categorized into six main types based on their physical characteristics: Apartment, House, Center building, Factory, High-rise building, and Others. The included buildings of the six categories are shown in Table 1.

**Table 1. Six building categories and corresponding buildings**

| Index | Categories | Buildings included |
| --- | --- | --- |
| a | Apartment | Multi-story residential buildings (lower than 7 layers or equal) |
| b | House | Individual houses, self-built houses |
| c | Center building | Commercial or mixed-use buildings |
| d | Factory | Industrial factories or storehouses |
| e | High-rise building | Tall commercial, residential buildings (higher than 7 layers), or towers |
| f | Others | Unique or specialized structures or Chinese classical architectures. |

## 2.2 Dataset Formation

### 2.2.1 Emerging city dataset: Zibo24

In this study, imagery was obtained from Google Earth, utilizing web scraping techniques to acquire remote sensing images of the Zibo region with a lens height of 500 meters. Subsequently, the images were segmented into square samples of equal area, resulting in the creation of the Zibo remote imaginary in the 2024 dataset (Zibo24). The detailed information and format of the Zibo24 dataset are illustrated in Table 2. This dataset comprises 149,420 high-resolution image samples, each with a resolution of 196 pixels per inch and representing a real-world area of 39,920 square meters. The scale of the images is 1:1110. However, as this data lacks segmentation labels, hyperparameter tuning, and model evaluation are not feasible. Therefore, a subset containing segmentation labels was constructed.

**Table 2. Format and information of Zibo24**

| Items | Description | values |
|---|---|---|
| Number of figures | Number of image samples in the dataset | 149420 |
| PPI | Pixels per inch for each standard image in the dataset | 196 |
| Scale | The ratio of building dimensions in the image to actual size | 1:1110 |
| Area per figure | The actual area represented by each standard image in the dataset | 39920m$^2$ |
| Hight | Height of the remote sensing lens | 500m |
| Labeled | With a label for Segmentation | No |

## 2.2.2 Subset with segment label

Hyperparameters and text prompt engineering settings significantly influence the segment accuracy of SAM, to find out the best hyperparameters and prompt text, a subset of Zibo satellite imagery with segment labels was collected. The details of this subset are shown in Table 3.

**Table 3. Format and information of Zibo24 subset**

| Items | Description | values |
|---|---|---|
| Number of figures | Number of image samples in the dataset | 100 |
| PPI | Pixels per inch for each standard image in the dataset | 196 |
| $S$ | The ratio of building dimensions in the image to actual size | 1:1110 |
| Area per figure | The actual area represented by each image in the dataset | 39920m$^2$ |
| Hight | Height of the remote sensing lens | 500m |
| Labeled | With a label for segmentation | Yes |
| Label categories | The number of label types | 6 |

As illustrated in Table 3, the Zibo24 subset consists of 100 human-labeled satellite imagery figures and the 100 figures are from Zibo24. They contain all six types of buildings in Zibo. Therefore, a comprehensive evaluation of different text prompts and hyperparameter settings can be established. The numbers of the six types of buildings and the shares of the total building roof area are shown in Figure 2 (a) and (b). In this subset, 42 individual buildings with Apartment labels with a share of 16.7% of the total building areas are organized. For House, Center building, Factory, High-rise building, and Others, the number of samples in each building category is 32, 11, 9, 15, and 12, each with a share of 11.1%, 23%, 20.6%, 8.7% and 19.8% of the total building areas. The diverse and balanced building categories within the Zibo24 subset provide a robust foundation for the selection and optimization of model hyperparameters.

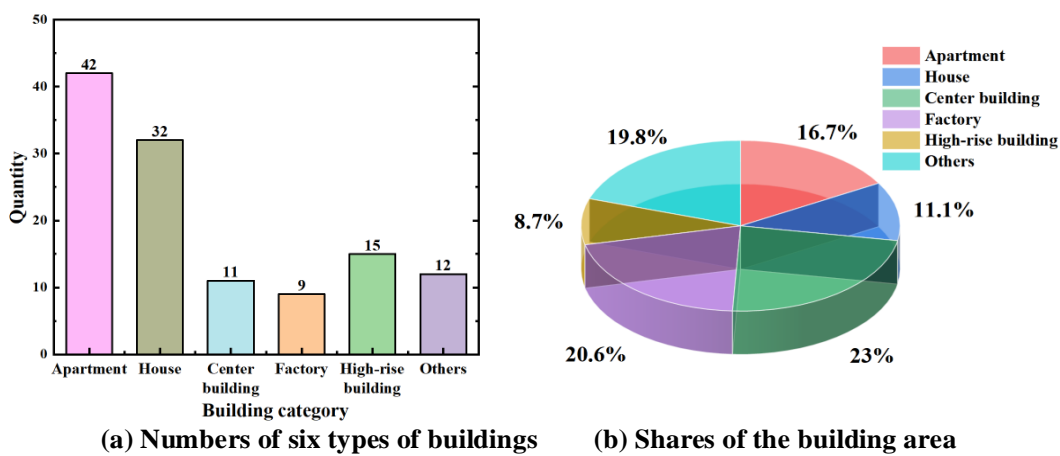

(a) Numbers of six types of buildings    (b) Shares of the building area

**Figure 2. Composition of the Zibo24 subset.**

# 3 Methodology

## 3.1 Building segmentation and area calculation

### 3.1.1 Building segmentation with SAM

To calculate the rooftop area of different building categories and assess the city-scale PV potential, a segment model, SAM, was utilized on the Zibo24 dataset. During the segmentation process, satellite images were fed into the SAM, and the model generated pixel-level masks that identify and isolate the building structures from the surrounding environment.

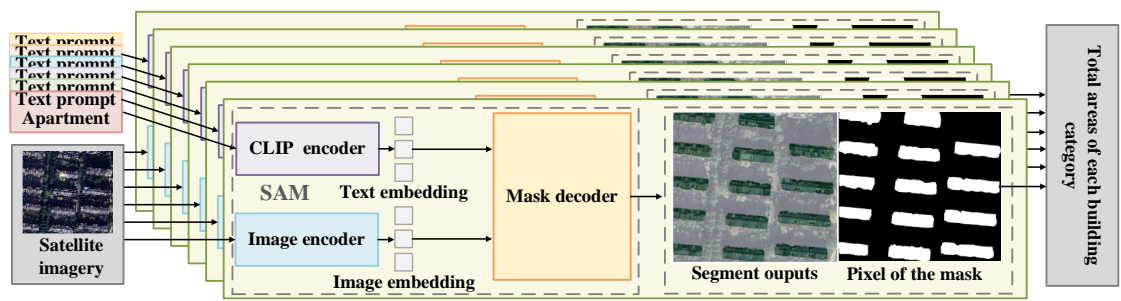

**Figure 3. SAM with different text prompt and area calculation of different building categories**

As illustrated in Figure 3, SAM is composed of a CLIP encoder, an Image encoder, and a Mask decoder. The CLIP encoder is utilized for understanding the text prompt and mapping natural language into word embedding. The Image encoder maps the satellite imagery into a high dimension space which is image embedding. Based on the word embedding and image embedding, the Mask decoder generates the segment mask. The segment mask from each image is guided by the corresponding text prompt. During this inference process, the segment mask for each satellite imagery is generated. To enhance the model's performance, the text prompts for the six building categories are separately given which means the segmentation processes repeat six times with six different text prompts. During the first five times, the names of the five building categories are utilized as the prompt text, and SAM generates the segment mask of the five different building categories. For the sixth time, the text prompt is "building", and SAM generates the mask of all the building roofs. To calculate the building area of the sixth category, "Other", the sum of the sixth group pixels minus the sum of the pixels from the other five groups is adopted, the details can be referenced from the next section.

To achieve accurate segmentation for each building type, this study investigated various text prompts on the labeled Zibo24 subset, and identified the optimal ones for each category. Defining the building categories as [building class], the text prompt phrases (TP1 to TP6) used to guide the SAM in segmenting different building categories are illustrated in Table 4.

**Table 4. Composite of a different text prompt**

| Index | Text prompt |
|-------|-------------|
| TP1 | [building class] |
| TP2 | [building class] + from satellite |
| TP3 | Roofs of + [building class] |
| TP4 | Roofs of + [building class] + from satellite |
| TP5 | Overhead shot of the + [building class] |
| TP6 | Many + [building class] + from satellite |

With the appropriate text prompts for each building category, buildings can be segmented using SAM. However, the segmentation accuracy is significantly influenced by the "box threshold" parameter, which determines the matching degree between the segmented object and the text prompt. A threshold that is too small may result in the segmentation of instances that do not belong to the target building category, while a threshold that is too large can lead to incomplete segmentation of the buildings. Therefore, this study utilized the Zibo24 subset to fine-tune this parameter. Following the segmentation process, the area of each identified building is calculated. This is achieved by leveraging the pixel count within each segmented building mask and applying the corresponding scale factor provided by the satellite imagery metadata. These area calculations provide quantitative data for different categories of buildings in Zibo.

### 3.1.2 Area calculation from masked pixel

The area calculation process utilizes the pixel counts within each segmented building mask generated by SAM and a scale factor derived from the satellite imagery metadata. Define $P_i$ as the set of pixels belonging to the mask of building category $i$, $P_b$ as the set of pixels in the mask of all the buildings, and $S$ as the scale factor converting pixels to real-world area units. The area of the six building categories, $A_i$, is calculated as follows:

$$A_i = P_i \times S, i \in (1,5)$$ (1)

$$A_6 = \left( P_b - P_1 - P_2 - P_3 - P_4 - P_5 \right) \times S$$ (2)

Utilizing this formula, the rooftop area of the six building categories can be calculated, and the areas for PV installation can be obtained.

### 3.1.3 Evaluation metrics

Semantic segmentation, as a well-established computer vision task, employs several evaluation metrics to assess the performance of segmentation models. Among these metrics, Pixel Accuracy (PA) and Intersection over Union (IoU) are two of the most commonly used. These metrics provide quantitative measures of the model's predictions aligned with the ground truth labels. For each building category, define the symbols as follows: *TP*: True Positives (pixels correctly classified); *TN*: True Negatives (pixels correctly classified as not belonging to this building category); *FP*: False Positives (pixels incorrectly classified as belonging to this building category while they actually belong to another category or the background), *FN*: False Negatives (pixels incorrectly classified as not belonging to this building category while they actually do). The metrics utilized in this paper are as follows:

- Pixel Accuracy (PA) calculates the proportion of pixels that are correctly classified. The calculation formula can be expressed as Equation (3):

$$PA = \frac{TP + TN}{TP + TN + FP + FN}$$ (3)

- Intersection over Union (IoU) calculates the proportion of the intersection area between the predicted segmentation and the ground truth segmentation to the union area. The calculation formula can be expressed as Equation (4):

$$IoU = \frac{TP}{TP + FP + FN} \tag{4}$$

These evaluation metrics offer valuable insights into the performance of SAM, serving as crucial benchmarks for optimizing hyperparameter selection and prompt engineering.

## 3.2 Multiple BIPV and PV models

This study analyzes the power generation potential of three BIPV types: Rooftop PV, Facade-integrated PV, and PV Windows, as illustrated in Figure 4. As PV systems in different locations receive varying amounts and angles of solar radiation, this paper constructs separate PV models for each BIPV type. Therefore, accurately city-scale PV power generation capabilities are assessed.

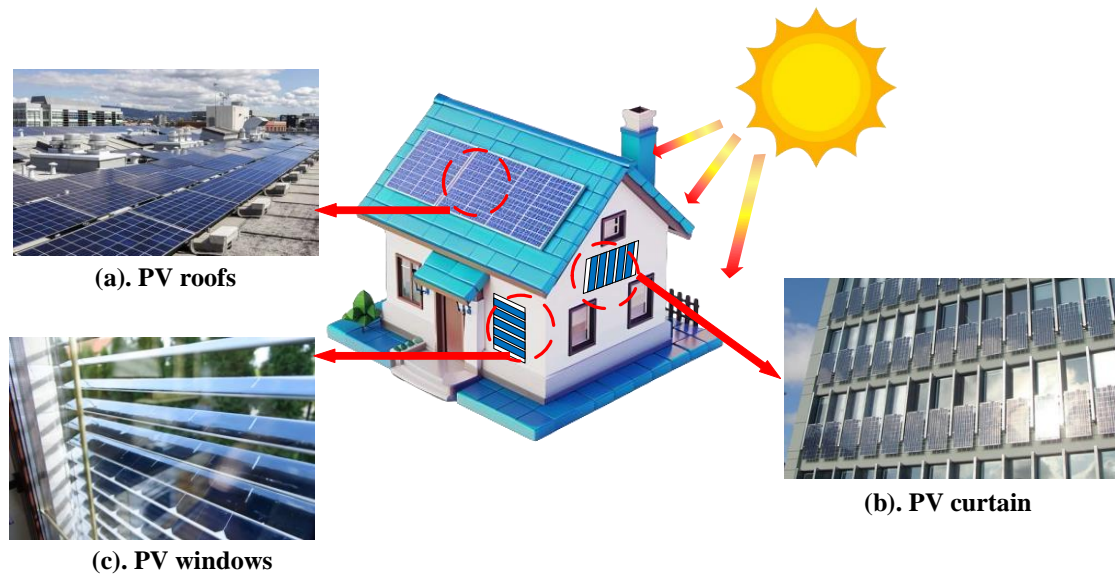

**(a). PV roofs**

**(b). PV curtain**

**(c). PV windows**

**Figure 4. Three types of BIPV were assessed in this work.**

### 3.2.1 Multiple BIPV

Rooftop PV, also known as rooftop solar power, is a mature and widely adopted form of BIPV [43,44]. Its power generation potential, economic benefits, and positive environmental impacts are well-established [45,46]. By replacing conventional roofing materials with solar panels, Rooftop PV transforms rooftops into power generators, harnessing sunlight to produce electricity.

Facade-integrated PV offers a cutting-edge solution for integrating solar power generation directly into a building's envelope [47]. This technology is particularly well-suited for high-rise buildings with large vertical surfaces, where Rooftop solar installations may be impractical.

PV windows represent an innovative advancement in BIPV, merging the functionality of traditional windows with the ability to generate clean electricity from sunlight [48]. These windows utilize transparent or semi-transparent solar cells embedded within the glass, allowing natural daylighting while simultaneously converting sunlight into usable energy. They offer a unique and aesthetically pleasing solution for incorporating solar power generation without compromising natural light or architectural design [49].

### 3.2.2 Area for BIPV installation

For the three BIPVs on the different building categories, the actual area available for PV installation (*AAPV*) is also different. Therefore, in this paper, an approximation conversion method is adopted. The formula is as follows:

$$AAPV_i = \sum_{BIPV} RA_i \cdot A \, / \, RA \cdot K_{mapping} \tag{5}$$

Where $AAPV_i$ is the *AAPV* from the building category, *i*. $A_i$ is the roof area of the building category, *i*. A/RA is the relation of the area between the area of the facade or windows and the rooftop, and the $K_{mapping}$ is the correction mapping factor, considering exclusions like existing structures, pip on the wall, or shading on the windows. For Facade-integrated PV and windows, surfaces that do not receive direct sunlight are unsuitable for PV installation. This factor is also considered in this mapping factor.

Summing up the actual area available for different BIPVs, the *AAPV* for each building category is calculated. The corresponding conversion factor can be referenced from Table 5.

**Table 5. Relevant parameters of a PV array in different buildings and BIPV**

| Buildings | Types of BIPV | | | | | |
| | Rooftop PV | | Facade-integrated PV | | PV windows | |
| | A/RA | $K_{mapping}$ | A/RA | $K_{mapping}$ | A/RA | $K_{mapping}$ |
|---|---|---|---|---|---|---|
| Apartment | 1 | 0.6 | 5.5 | 0.35 | 0.18 | 0.10 |
| House | 1 | 0.7 | 1.4 | 0.30 | 0.12 | 0.18 |
| Center building | 1 | 0.65 | 1.8 | 0.35 | 0.25 | 0.12 |
| Factory | 1 | 0.8 | 1.3 | 0.40 | 0.08 | 0.06 |
| High-rise building | 1 | 0.5 | 12.1 | 0.24 | 0.35 | 0.18 |
| Others | 1 | 0.4 | 1.1 | 0.15 | 0.20 | 0.03 |

### 3.2.3 Potential of PV power generation

To estimate the electricity power generation of different BIPVs, this study leverages the Pvlib Python package [50]. It is renowned for its reliable, interoperable, and benchmark implementations of PV system models. Pvlib is capable of modeling tilt angle, and module parameters from data affecting them, such as climate conditions. This makes the setting of parameters for PV modules, inverters, and other system components easier. In this study, the parameters of the selected PV panel are shown in Table 6, and the other parameters of the climate conditions can be referenced in the supplementary data. The three types of BIPVs in the six building categories have different installation locations. This leads to different shading patterns, irradiance, and solar angles. Therefore, this study creates 18 separate PV models for each BIPV type in each building category.

**Table 6. Parameters of the PV panel for Zibo**

| Parameters | Values |
|---|---|
| Length | 2094mm |
| Width | 1038mm |
| Depth | 35mm |
| Weight | 27.5kg |
| Installation Area | 2.23m$^2$ |
| Maximum Power ($P_{max}$/W) | 46.3 |
| Voltage at Maximum Power ($V_{mp}$/V) | 41.0 |
| Current at Maximum Power ($I_{mp}$/A) | 11.45 |
| Operating Temperature ($T_{OT}$) | -40°C~+85°C |

As illustrated in Table 6. This panel boasts a maximum power output of 46.3 W, it demonstrates efficient energy generation capabilities. The voltage at maximum power is 41.0 V, while the corresponding current stands at 11.45 A. Furthermore, the PV panel exhibits a wide operating temperature range, functioning effectively from a frigid -40°C up to a scorching 85°C, ensuring reliable performance for the yearly power generation in Zibo.

## 3.3 Economic and environmental metrics for PV

### 3.3.1 Levelized cost of electricity

To assess the economic potential of different BIPVs on different buildings, the levelized cost of electricity (LCOE) is utilized [51]. LCOE represents the average net present cost of electricity generation over the lifetime of the PV system, encompassing all costs associated with installation, operation, maintenance, and decommissioning, discounted to present value [52]. To calculate the LCOE, the annual PV electricity generation at $t$ year, $E_t$ (kWh), and the annual total cost, $C_t$ (CNY) should be first calculated. In this study, a deep learning-based model [53] is utilized to forecast the annual power generation, $E_t$, from 2023 to 2047 in Zibo. This model could predict the annual power generation from the history data. To calculate $C_t$, the relative parameters are defined as follows: $C_{start}$ is the initial investment for purchasing and installing PV panels, inverters, and other system components, $C_{O\&M}$ is the fee for the operating and maintenance of the PV system, and $C_{rent}$ is the cost of renting rooftops, walls, and windows. These costs are obtained from the local trading website, hardware provider, and reference [54]. Therefore, the cost of the BIPV system at $t$ year can be calculated as follow:

$$C_t = \begin{cases} C_{start} + C_{O\&M} + C_{rent}, t = 1 \\ C_{O\&M} + C_{rent}, t > 1 \end{cases} \quad (6)$$

And the LCOE (CNY/kWh) can be calculated as follow:

$$LCOE = \frac{\sum_{t=1}^{n} \dfrac{C_t}{(1+r)^t} \times Capacity}{\sum_{t=1}^{n} \dfrac{E_t}{(1+r)^t}} \quad (7)$$

Where $r$ is the discount rate which is 0.05 in this work, $n$ is 25 and *Capacity* is the installed PV capacity (W). By calculating the LCOE for different building categories with different BIPVs, we can compare their economic feasibility and identify the most cost-effective solutions for widespread PV panel deployment in Zibo. Lower LCOE values indicate greater economic viability and competitiveness compared to traditional electricity sources. This can inform targeted deployment strategies and policy recommendations to prompt and optimize the BIPV installation in emerging cities.

### 3.3.2 Carbon emission

Carbon emission reduction calculation revels the environmental impact of the different BIPV for an emerging city. This evaluation calculates equivalent electricity generation from fossil fuel power plants displaced by PV systems based on a relevant carbon emission factor [55]. The formula for carbon emission reduction (CER) is expressed as Equation (8):

$$CER = E_{pv} \cdot EF \tag{8}$$

where $E_{pv}$ represents the total electricity generated by PV systems (kWh) and $EF$ denotes the carbon emission factor of the displaced fossil fuel power plant (kg $CO_2$/kWh). In this paper, the $EF$ is obtained from the Ministry of Ecology and Environment of the People's Republic of China, which is 0.6838 kg $CO_2$/kWh. Through this analysis, the environmental benefits of transitioning towards solar energy in Zibo can be quantified, supporting policy decisions and sustainable development initiatives. The emission reduction potential across various building categories and BIPV further informs targeted PV deployment strategies for maximizing environmental impact while considering economic feasibility and urban development goals.

# 4 Results and Discussions

## 4.1 SAM results

### 4.1.1 Prompt engineering of SAM

Utilizing the SAM for processing satellite imagery allows for the semantic segmentation of different building types. However, as SAM is a general-purpose semantic segmentation model, it requires specific text prompts to guide its output toward the desired segmentation targets. The choice of text prompt significantly impacts the segmentation results. Therefore, based on the Zibo24 subset and the text prompts in Table 4, the ablation study of prompt tokens was conducted. By comparing the segmentation outputs obtained with these different text prompts, this study identified the optimal prompts for each building category. The results of IoU with different text prompts are shown in Figure 5. The prompt "Roofs of [building class] from satellite" consistently yielded the best or near-best performance for most building types. Specifically, this prompt achieved IoU values of 36.2, 28.6, and 30.2 for the categories of Apartment, House, and Center building. Furthermore, emphasizing the presence of multiple buildings with the prompt "Many [building class] from satellite" enhanced results for the building category of Factory, achieving a IoU of 31.2. Similarly, highlighting the overhead perspective with the prompt "Overhead shot of the [building class]" proved beneficial for the High-rise building category, resulting in a IoU of 31.1. During this process, the best prompt text for the segmentation of each building category is determined.

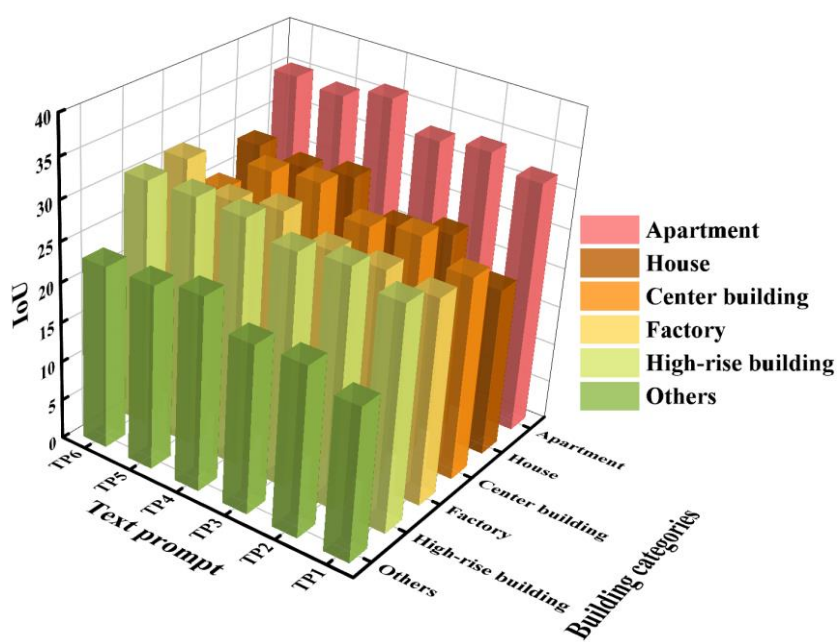

**Figure 5. IoU of SAM on Zibo24 with different text prompts.**

To provide a clearer visual comparison of the impact of different text prompts on building segmentation, the segment results for the "Apartment" category are illustrated in Figure 6. It is evident that utilizing "TP4: Roofs of Apartment from satellite" as the text prompt yields the most accurate results. With this prompt, the model accurately segments buildings while minimizing extraneous segmentation errors. In contrast, the first three text prompts (TP1 to TP3) exhibit shortcomings, including misidentifying building shadows as rooftops and failing to capture all building structures. As demonstrated in Figure 6 (e) and (f), employing TP5 and TP6 as modifiers improves performance compared to using "Apartment" alone. However, upon analyzing the segmentation results, it becomes apparent that these prompts fail to comprehensively segment all the buildings. Therefore, the text structure of TP4 proves to be the most effective for the segmentation of Apartment. And for the other five building categories, it is similar.

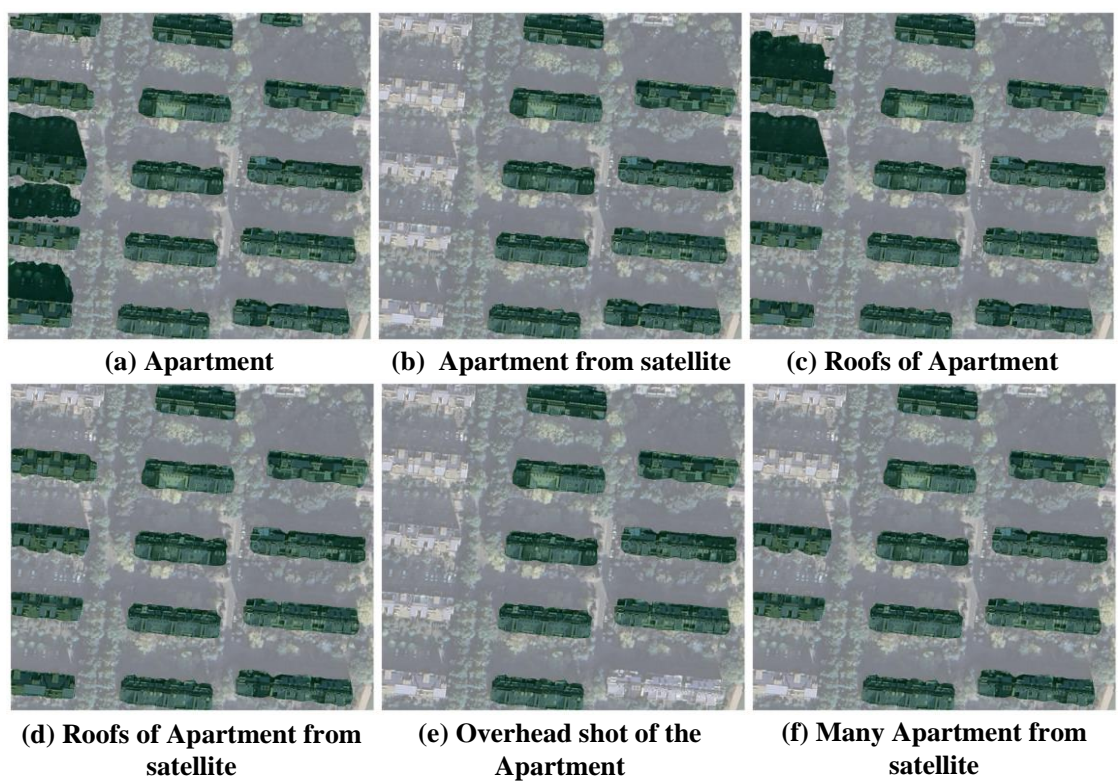

**(a) Apartment**     **(b) Apartment from satellite**     **(c) Roofs of Apartment**

**(d) Roofs of Apartment from satellite**     **(e) Overhead shot of the Apartment**     **(f) Many Apartment from satellite**

**Figure 6. Building segmentation results from SAM with different text prompts.**

## 4.1.2 Hyperparameter optimization

Hyperparameter, box threshold exhibits the most significant influence on the segment accuracy. Therefore, this study employs a grid search approach to optimize the hyperparameter for each building category, with the corresponding best text prompt from the Zibo24 subset. The results of this optimization process are presented in Figure 7, and segment results with different box thresholds are shown in Figure 8.

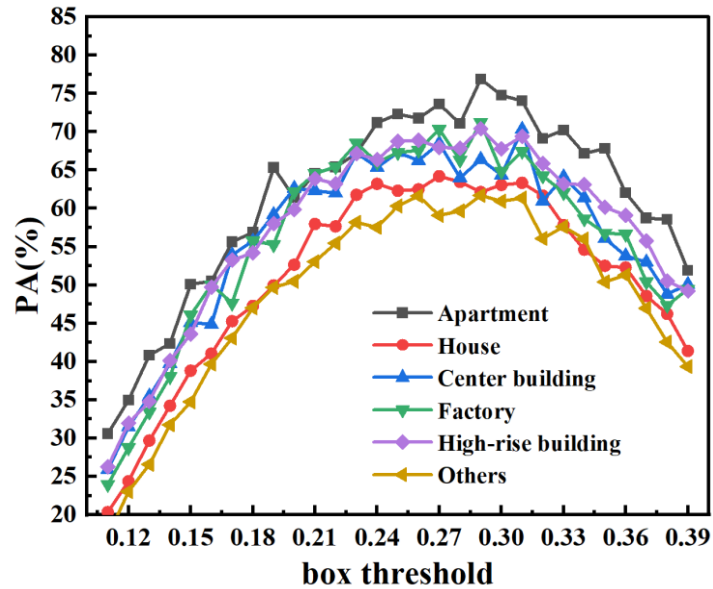

**Figure 7. SAM with different hyperparameter settings.**

As shown in Figure 7, the box threshold has a significant impact on the results of building segmentation. When the box threshold is low, the PA is relatively low, because the pixels that do not belong to the building are disturbed by the segment result, as shown in Figure 8 (a). When the box threshold is higher, the PA is also bigger, for the less pixels are miss segmented, as shown in Figure 8 (b) and (c), but there still are some pixels miss segmented into the building roofs. When the box threshold is around 0.25, the optimal box threshold can be found for the different building categories. However, when the threshold is too big, the PA becomes lower, because the building instances cannot be segmented, as shown in Figure 8 (e) and (f).

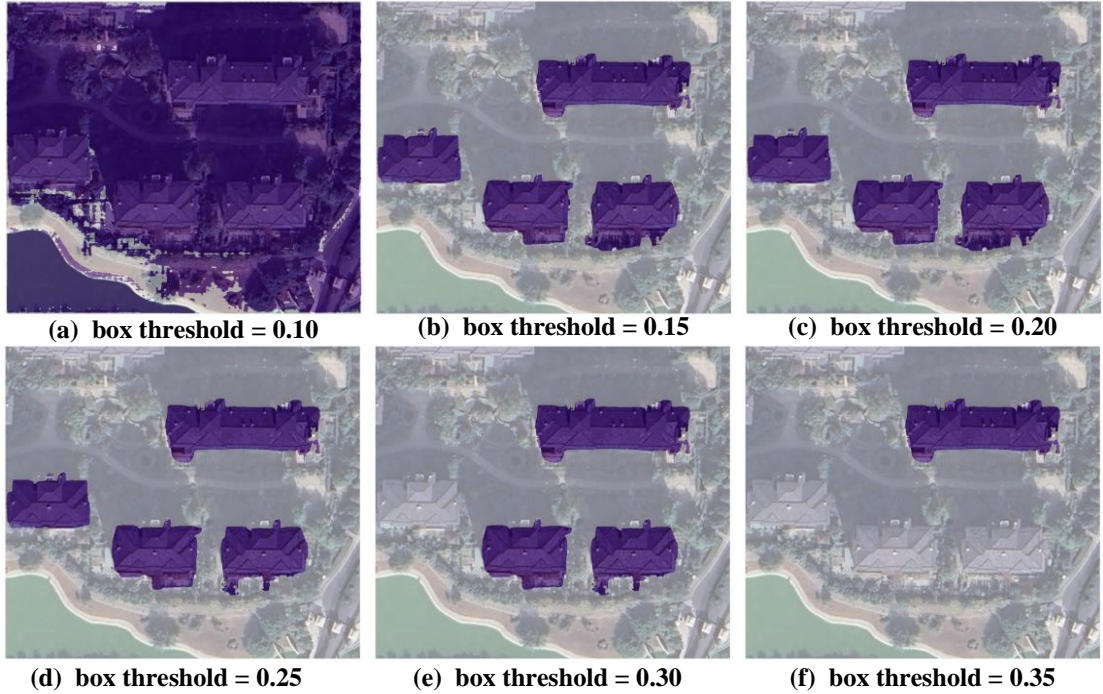

(a) box threshold = 0.10     (b) box threshold = 0.15     (c) box threshold = 0.20

(d) box threshold = 0.25     (e) box threshold = 0.30     (f) box threshold = 0.35

**Figure 8. Building segmentation results from SAM with different box threshold**

### 4.1.3 Segment results and area calculation

Based on the Zibo24 subset, the optimal model hyperparameters were obtained. To calculate the total rooftop area for each of the six building categories in Zibo, the entire Zibo24 dataset was utilized. By summing the pixel count within the mask layer of the semantic segmentation output and multiplying it with the scale factor, the total area for each building category was calculated, the segmentation results of six building types can be referenced from Figure 9, and total areas can are illustrated in Table 7. In Figure 9, with the optimal text prompt and best box threshold for each building category, Apartment, House, Center building, Factory, High-rise building, and Others can be separately segmented.

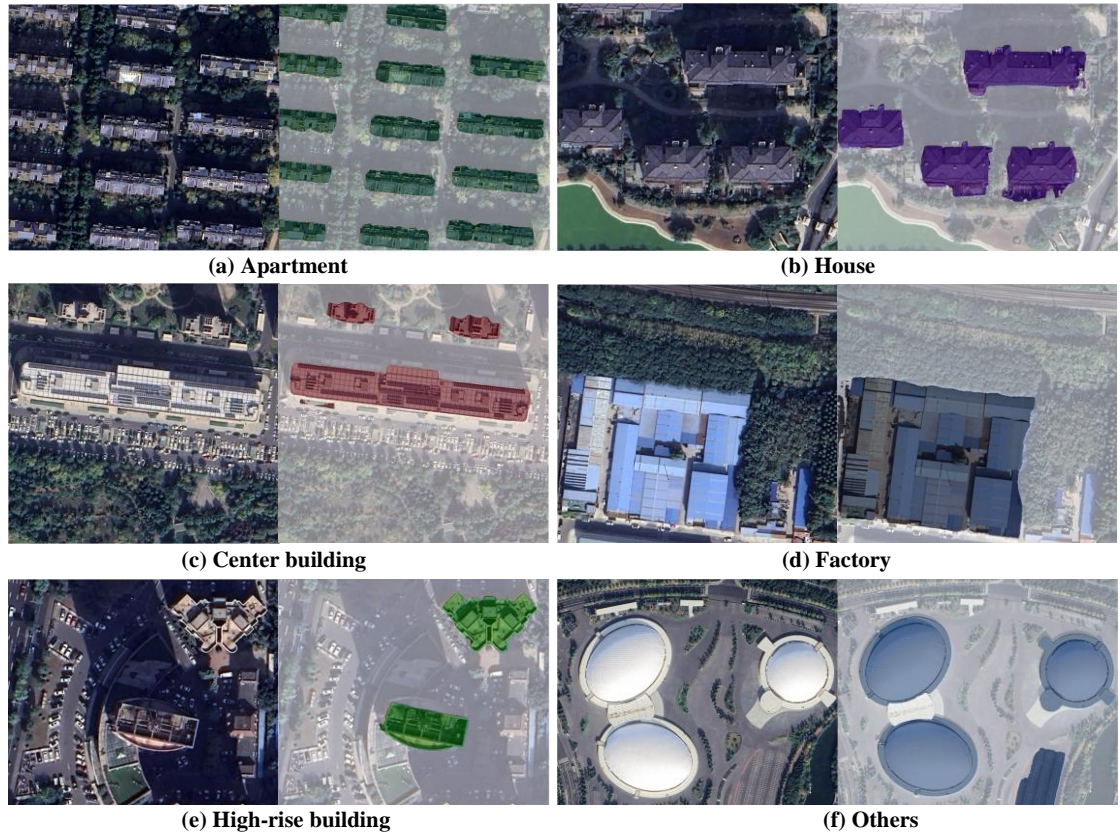

(a) Apartment        (b) House

(c) Center building        (d) Factory

(e) High-rise building        (f) Others

**Figure 9. Segment results of different building types in Zibo.**

As illustrated in Table 7, the building roof areas in Zibo are calculated through the segmented pixel with the mapping of scales. The Factory category, with a total roof area of 82,321,505 square meters (sqm), clearly surpasses any other category, highlighting the city's strong industrial presence, and the potential of BIPV installation for industrial usage. Residential areas, comprising the categories of Apartment (56,194,238 sqm) and House (32,877,425 sqm), collectively occupy a substantial portion of land, suggesting significant parts for residual PV. Notably, the category of Center building, covering 44,583,152 sqm, holds more area than the High-rise building category (9,843,321 sqm). And for the "Other" category, the roof area is 86,554,648 sqm. Based on calculations of this area, the installation potential of different BIPV types can be further assessed.

**Table 7. Total areas of different types of building roofs**

| Buildings | Total area of the rooftop (m$^2$) |
| --- | --- |
| Apartment | 56,194,238 |
| House | 32,877,425 |
| Center building | 44,583,152 |
| Factory | 82,321,505 |
| High-rise building | 9,843,321 |
| Others | 86,554,648 |

## 4.2 PV potential of Zibo

### 4.2.1 BIPV installation potential

Taking into account the physical dimensions of the selected PV panels and the effective installation areas for Rooftop PV, Facade-integrated PV, and PV windows, we calculated the potential number of PV panels for each BIPV system across various building categories, as detailed in Table 8. Figure 10 provides a visual representation of installation potential for different BIPVs across various building categories. Analysis of Zibo's BIPV potential reveals a clear dominance of Rooftop PV installations, with Factory offering the highest capacity at 30.2 million PV panels, followed by Others and Apartment with 15.9 and 15.5 million. But for the High-rise building, the available installation number of PV panels on the rooftop is only 2.2 million. Facade-integrated PV presents a strong alternative, particularly for high-rise buildings with a potential of 13.1 million panels. And for Apartment, it is 49.7 million which is the highest among all the building categories. PV windows offer the least capacity across all the BIPVs. The PV windows installation potential of the Center building is the highest (0.6 million), and for Factory the potential is lowest among all the building categories. Notably, residential buildings exhibit significant potential for solar energy generation, with Apartment and House categories collectively accommodating over 26 million PV panels on their rooftops, 56 million PV panels on the Facade, and 0.7 million on their window. Center building also demonstrates strong potential for Rooftop PV and Facade-integrated PV, with capacities exceeding 13.3 million and 12.9 million panels respectively. These findings highlight the diverse opportunities for BIPV integration across Zibo's urban landscape and underscore the potential of building-scale solar power generation in achieving a more sustainable future.

**Table 8. Total numbers of PV panels on different building categories**

| Buildings | Types of BIPV | | |
|---|---|---|---|
| | Rooftop PV | Facade-integrated PV | PV windows |
| Apartment | 15,512,043 | 49,767,805 | 465,361 |
| House | 10,588,191 | 6,352,915 | 326,721 |
| Center building | 13,332,454 | 12,922,224 | 615,343 |
| Factory | 30,299,067 | 19,694,393 | 181,794 |
| High-rise building | 2,264,319 | 13,151,164 | 285,304 |
| Others | 15,928,554 | 6,570,528 | 238,928 |

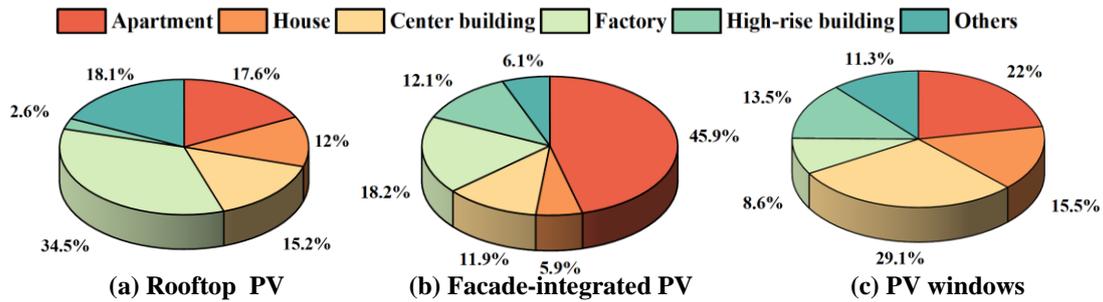

**Figure 10. Comparison of installation potential for different BIPVs across various building categories.**

## 4.2.2 Electricity generation potential of different BIPV

Leveraging regional meteorological data from Zibo in 2022, we calculated the estimated power generation potential for three types of BIPV across six building categories. The results of this analysis are visually presented in Figure 11. Rooftop PV consistently demonstrates a strong capacity for power generation across all building categories, with the Factory category leading the way at 18.18 TWh annually, followed by Apartment at 9.31 TWh and Others at 9.56 TWh. This aligns with the large roof areas available in these building types. While Facade-integrated PV exhibits lower overall potential compared to Rooftop PV, they present a compelling alternative for specific categories. Apartment, with extensive facades, show a remarkably high potential for Facade-integrated PV installations, generating an estimated 17.92 TWh annually, almost double the energy output of their Rooftop PV and surpassing the generation capacity of any other building category using Facade-integrated PV. High-rise building also demonstrates a preference for Facade-integrated PVs, generating 4.34 TWh annually, over three times the output of their Rooftop PV. In contrast, PV windows exhibit the lowest potential across all categories, contributing minimally to the overall energy generation landscape. These findings underscore the critical role of building type and facade area in determining the optimal BIPV system for maximizing renewable energy generation in Zibo.

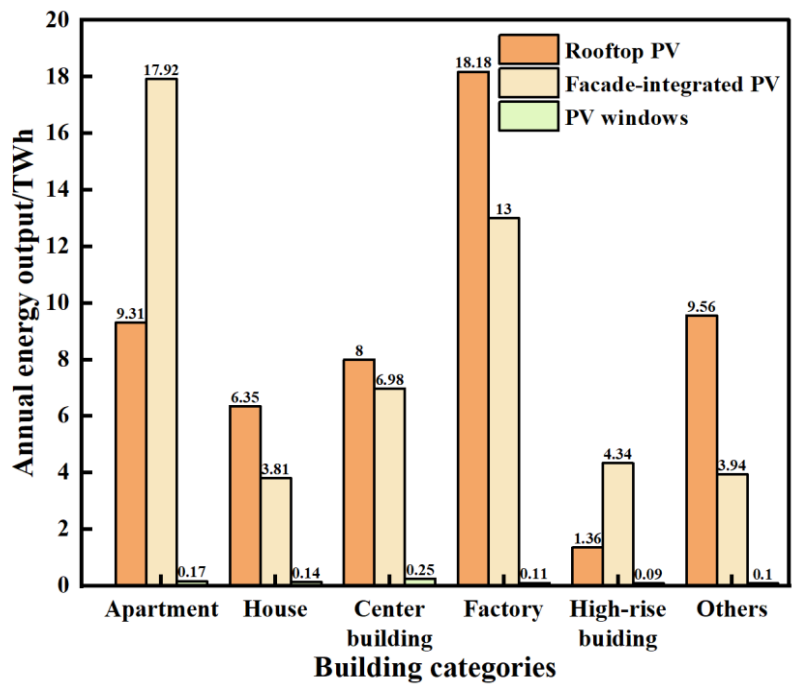

Figure 11. The electricity power generation potential of different BIPVs in Zibo.

### 4.2.3 Powering the emerging city solely with BIPV

Utilizing BIPV models and meteorological data collected from 2013 to 2022 in Zibo, the BIPV power generation over these ten years was computed. This analysis, depicted in Figure 12, further incorporates annual electricity consumption data for Zibo obtained from the Shandong Provincial Bureau of Statistics. Zibo boasts a remarkable potential for solar energy generation that consistently surpasses its electricity consumption, even with the city's growing energy demands. Throughout the past decade, from 2013 to 2022, the estimated annual electricity generation from BIPV consistently outpaced total electricity consumption, often by a factor of 2.5 or more. In 2022, for example, the BIPV potential reached 103.59 TWh, dwarfing the total electricity consumption of 41.63 TWh. While Zibo's electricity needs have steadily risen by nearly 27% during this period (rising from 32.84 TWh in 2013 to 41.63 TWh in 2022), the stable (fluctuating only within a range of approximately 4 TWh yearly for climate factor) and abundant solar resource ensures a surplus of clean energy generation potential. This presents a compelling opportunity for Zibo to transition towards a sustainable and self-sufficient energy system, reducing reliance on fossil fuels, improving air quality, and fostering economic growth through green energy investments. By harnessing the power of the sun, Zibo can illuminate a path toward a brighter and more sustainable future.

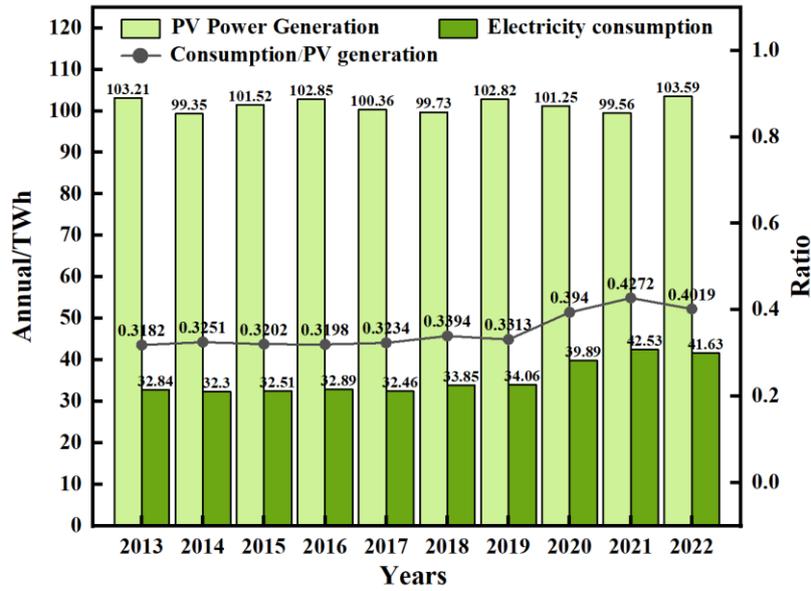

**Figure 12. Annual Power generation from different BIPV and the electricity consumption in Zibo.**

## 4.3 Economic and environmental analysis

### 4.3.1 Economic analysis (LCOE)

To evaluate the economic potential for the different BIPVs, LCOE analyses are adopted. The results and the total costs of the different PV systems are illustrated in Figure 13. Examining the LCOE data for various BIPV systems across different building types in Zibo reveals crucial insights into the economic potential of solar energy. Rooftop PV is a cost-effective choice. Rooftop PV systems exhibit remarkable consistency in LCOE figures across diverse building categories. Factory benefits from the lowest LCOE at 0.18 CNY/kWh (0.0252 $/kWh, exchange rate=0.14), followed closely by House and Center building at 0.19 CNY/kWh (0.0266 $/kWh). Apartment also shows a competitive LCOE of 0.19 CNY/kWh (0.0266 $/kWh), while high-rise buildings experience a slightly higher LCOE of 0.20 CNY/kWh (0.028 $/kWh) due to the increased complexity of installations on taller structures.

Facade-integrated PV systems are also potential for Zibo. Facade-integrated PV systems present a more nuanced economic picture, with LCOE figures influenced by building characteristics and space rent costs. House demonstrates the most attractive LCOE at 0.21 CNY/kWh (0.0294 $/kWh), followed by Factory at 0.17 CNY/kWh (0.0238 $/kWh). However, the LCOE for Apartment increases to 0.34 CNY/kWh (0.0476 $/kWh), and for High-rise building, it reaches 0.37 CNY/kWh (0.0581 $/kWh), primarily due to the higher space rent of 0.5 CNY/kWh (0.07 $/kWh) associated with utilizing building facades. Center building and Others building categories fall in between, with LCOEs of 0.25 CNY/kWh (0.035 $/kWh) and 0.23 CNY/kWh (0.0322 $/kWh), respectively.

PV Window systems are a good solution for specific situations. Factory achieved lowest LCOE at 0.20 CNY/kWh (0.028 $/kWh) because of the reduction of engineering costs and lower space rent. In contrast, the LCOEs for Apartment and High-rise building rise to 0.39 CNY/kWh (0.0546 $/kWh) and 0.41 CNY/kWh (0.0574 $/kWh) respectively, primarily driven by the higher space rents of 1.5 CNY/W (0.21 $/W) and 0.75 CNY/W (0.105 $/W) for utilizing window areas.

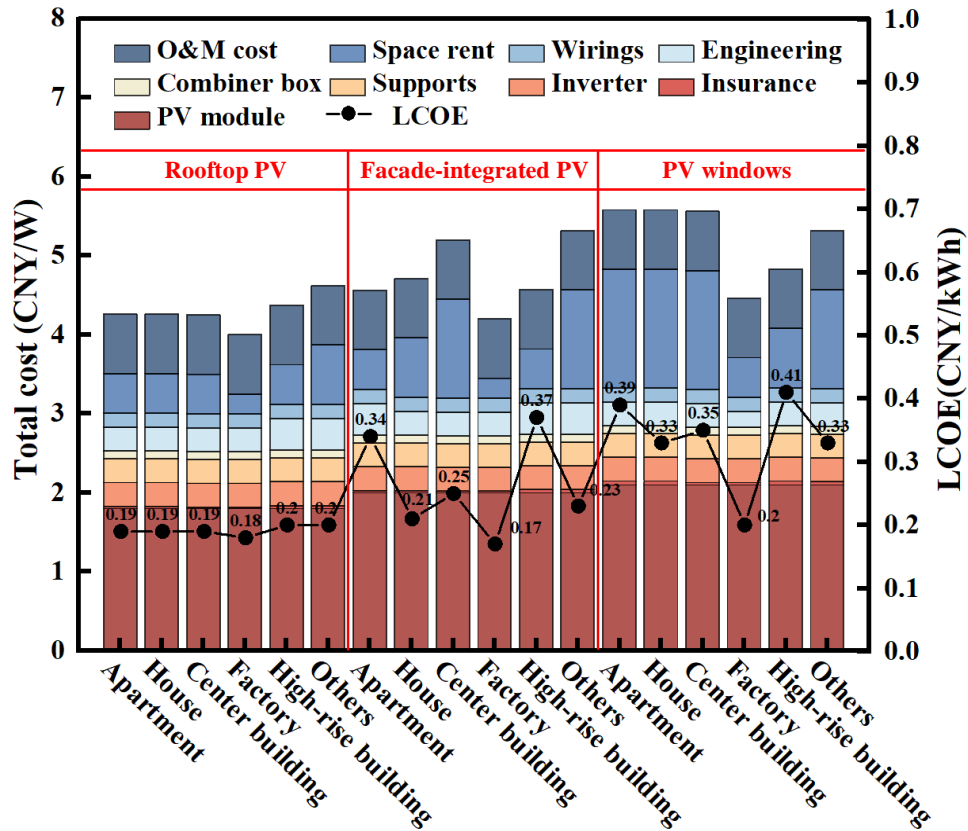

**Figure 13. Cost of different BIPV on different building categories and the LCOE.**

All in all, Rooftop PV maintains a clear economic advantage across all building categories. Facade-integrated PV has potential for buildings with lower space rent and ample facade area, such as House and Factory, but becomes less attractive for taller structures for the higher space rent and installation cost or those with limited facade space. PV windows present a specialized solution with competitive LCOE in specific scenarios, such as factories, but their higher initial costs and space rent requirements can limit their broader applicability. Therefore, considering the economic feasibility and specific building characteristics, stakeholders can make informed choices regarding the optimal BIPV technology for maximizing both energy generation and financial returns.

### 4.3.2 Environmental analysis

To calculate the annual carbon emission reduction, this study used carbon emission factors and converted the power generation of different BIPV systems installed on different building categories. The reduction of carbon emissions is illustrated in Table 9.

**Table 9. Yearly reduction of carbon emission from different BIPV on different building categories in Zibo ($\times 10^7$ T $CO_2$)**

| Buildings | Types of BIPV | | |
|---|---|---|---|
| | Rooftop PV | Facade-integrated PV | PV windows |
| Apartment | 0.636,6 | 1.225,4 | 0.011,6 |
| House | 0.434,2 | 0.260,5 | 0.009,6 |
| Center building | 0.547 | 0.477,3 | 0.017,1 |
| Factory | 1.243,1 | 0.888,9 | 0.007,5 |
| High-rise building | 0.093 | 0.296,8 | 0.006,2 |
| Others | 0.653,7 | 0.269,4 | 0.006,8 |

The carbon emission reduction potential of different BIPVs reveals a promising pathway toward a more sustainable future. Rooftop PV is the most significant participant in carbon reduction, especially for the buildings with large roof areas, such as Factory, and it achieved a reduction of $1.2431 \times 10^7$ T $CO_2$ in one year. Apartment also demonstrates substantial potential with Rooftop PV, with a reduction of $0.6366 \times 10^{17}$ T $CO_2$, highlighting the impact of residential buildings on carbon reduction. Facade-integrated PV is powerful for buildings with limited roof space. For example, Apartment achieves an even greater emission reduction of $1.2254 \times 10^7$ T $CO_2$ using Facade-integrated PV. This shown the immense potential of utilizing vertical surfaces. The PV windows also shown its carbon reduction potential in High-rise building and Center building with abundant window surfaces. The yearly carbon reduction of all the BIPVs in Zibo is $7.0847 \times 10^7$ T $CO_2$. In conclusion, this multifaceted approach, incorporating Rooftop PV, Facade-integrated PV, and PV windows, unlocks a vast potential for carbon emission reduction in Zibo's building sector, paving the way for a greener and more sustainable urban environment.

# 5 Conclusion

This study proposed the first model for evaluating BIPV potential in emerging cities using the SAM and remote sensing data. The model, SolarSAM, is applied to Zibo, a rapidly developing city in China, to quantify the available installation area of Rooftop PV, Facade-integrated PV, and PV windows across six building categories. Separate PV models were developed for each BIPV type, accounting for distinct installation and application scenarios. This comprehensive approach enabled a thorough comparison of different BIPV options, considering their potential for city-wide PV energy self-sufficiency, economic benefits, and environmental impact. The key findings are as follows:

1. Based on the selected text prompt, SolarSAM has proven to be an effective model for segmenting different building categories in remote sensing images, enabling the calculation of potential BIPV installation areas solely from remote sensing data.

2. SolarSAM reveals the BIPV potential in Zibo, with an annual power generation capacity of approximately 101 TWh – exceeding the city's total electricity consumption by a factor of 2.5. This highlights the remarkable potential for a clean energy future in Zibo and other emerging cities.

3. Rooftop PV demonstrates the highest potential across various building categories. However, for Apartment and High-rise buildings, Facade-integrated PV offers greater installation potential. Among all BIPVs, the installation potential of PV windows remains the lowest.

4. Rooftop PV emerges as the most cost-effective option across all building categories, with a LCOE below 0.20 CNY/kWh (0.028 $/kWh). Facade-integrated PV proves competitive for Factory and House categories.

5. BIPV systems in Zibo have the potential to significantly reduce carbon emissions, achieving an estimated annual reduction of $7.0847 \times 10^7$ T $CO_2$. This substantial contribution to a cleaner and more sustainable urban environment is particularly significant for emerging cities like Zibo, which are undergoing rapid urbanization and industrial development.

In the future, advancements in CV model performance will further enhance building segmentation accuracy, and this will lead to improved accuracy in comprehensive city-wide BIPV potential assessments. This research contributes significantly to the development of sustainable urban environments in emerging cities. By providing a robust and adaptable framework for emerging city BIPV potential assessment, it supports the implementation of clean energy solutions and contributes to the realization of China's carbon neutrality goals. Our findings provide valuable insights for policymakers, urban planners, and industry stakeholders, enabling informed decision-making and driving the adoption of BIPV technologies.

**Data Availability**

The source code can be referenced from https://github.com/REAILAB/SolarSAM.git.

**Reference**

[1] Q. Hassan, P. Viktor, T. J. Al-Musawi, B. Mahmood Ali, S. Algburi, H.M. Alzoubi, A. Khudhair Al-Jiboory, A. Zuhair Sameen, H.M. Salman, M. Jaszczur, The renewable energy role in the global energy Transformations, Renewable Energy Focus 48 (2024) 100545. https://doi.org/10.1016/j.ref.2024.100545.

[2] P. Choudhary, R.K. Srivastava, Sustainability perspectives- a review for solar photovoltaic trends and growth opportunities, Journal of Cleaner Production 227 (2019) 589–612. https://doi.org/10.1016/j.jclepro.2019.04.107.

[3] U. Perwez, K. Shono, Y. Yamaguchi, Y. Shimoda, Multi-scale UBEM-BIPV coupled approach for the assessment of carbon neutrality of commercial building stock, Energy and Buildings 291 (2023) 113086. https://doi.org/10.1016/j.enbuild.2023.113086.

[4] K. Bódis, I. Kougias, A. Jäger-Waldau, N. Taylor, S. Szabó, A high-resolution geospatial assessment of the rooftop solar photovoltaic potential in the European Union, Renewable and Sustainable Energy Reviews 114 (2019) 109309. https://doi.org/10.1016/j.rser.2019.109309.

[5] M. Manni, M. Formolli, A. Boccalatte, S. Croce, G. Desthieux, C. Hachem-Vermette, J. Kanters, C. Ménézo, M. Snow, M. Thebault, M. Wall, G. Lobaccaro, Ten questions concerning planning and design strategies for solar neighborhoods, Building and Environment 246 (2023) 110946. https://doi.org/10.1016/j.buildenv.2023.110946.

[6] H. Sun, C.K. Heng, S.E.R. Tay, T. Chen, T. Reindl, Comprehensive feasibility assessment of building integrated photovoltaics (BIPV) on building surfaces in high-density urban environments, Solar Energy 225 (2021) 734–746. https://doi.org/10.1016/j.solener.2021.07.060.

[7] T.E. Kuhn, C. Erban, M. Heinrich, J. Eisenlohr, F. Ensslen, D.H. Neuhaus, Review of technological design options for building integrated photovoltaics (BIPV), Energy and Buildings 231 (2021) 110381. https://doi.org/10.1016/j.enbuild.2020.110381.

[8] T. Yang, A.K. Athienitis, A review of research and developments of building-integrated photovoltaic/thermal (BIPV/T) systems, Renewable and Sustainable Energy Reviews 66 (2016) 886–912. https://doi.org/10.1016/j.rser.2016.07.011.

[9] M. Shafique, X. Luo, J. Zuo, Photovoltaic-green roofs: A review of benefits, limitations, and trends, Solar Energy 202 (2020) 485–497. https://doi.org/10.1016/j.solener.2020.02.101.

[10] Y. Tang, J. Ji, H. Xie, C. Zhang, X. Tian, Single- and double-inlet PV curtain wall systems using novel heat recovery technique for PV cooling, fresh and supply air handling: Design and performance assessment, Energy 282 (2023) 128797. https://doi.org/10.1016/j.energy.2023.128797.

[11] D. Liu, Y. Sun, R. Wilson, Y. Wu, Comprehensive evaluation of window-integrated semi-transparent PV for building daylight performance, Renewable Energy 145 (2020) 1399–1411. https://doi.org/10.1016/j.renene.2019.04.167.

[12] A.K. Shukla, K. Sudhakar, P. Baredar, A comprehensive review on design of building integrated photovoltaic system, Energy and Buildings 128 (2016) 99–110. https://doi.org/10.1016/j.enbuild.2016.06.077.

[13] M. Panagiotidou, M.C. Brito, K. Hamza, J.J. Jasieniak, J. Zhou, Prospects of photovoltaic rooftops, walls and windows at a city to building scale, Solar Energy 230 (2021) 675–687. https://doi.org/10.1016/j.solener.2021.10.060.

[14] V. Kapsalis, C. Maduta, N. Skandalos, M. Wang, S.S. Bhuvad, D. D'Agostino, T.

Ma, U. Raj, D. Parker, J. Peng, D. Karamanis, Critical assessment of large-scale rooftop photovoltaics deployment in the global urban environment, Renewable and Sustainable Energy Reviews 189 (2024) 114005. https://doi.org/10.1016/j.rser.2023.114005.

[15] G. Yu, H. Yang, Z. Yan, M. Kyeredey Ansah, A review of designs and performance of façade-based building integrated photovoltaic-thermal (BIPVT) systems, Applied Thermal Engineering 182 (2021) 116081. https://doi.org/10.1016/j.applthermaleng.2020.116081.

[16] G. Yu, H. Yang, D. Luo, X. Cheng, M.K. Ansah, A review on developments and researches of building integrated photovoltaic (BIPV) windows and shading blinds, Renewable and Sustainable Energy Reviews 149 (2021) 111355. https://doi.org/10.1016/j.rser.2021.111355.

[17] L. Pratt, D. Govender, R. Klein, Defect detection and quantification in electroluminescence images of solar PV modules using U-net semantic segmentation, Renewable Energy 178 (2021) 1211–1222. https://doi.org/10.1016/j.renene.2021.06.086.

[18] A. Mellit, S. Kalogirou, Assessment of machine learning and ensemble methods for fault diagnosis of photovoltaic systems, Renewable Energy 184 (2022) 1074–1090. https://doi.org/10.1016/j.renene.2021.11.125.

[19] D. Assouline, N. Mohajeri, J.-L. Scartezzini, Quantifying rooftop photovoltaic solar energy potential: A machine learning approach, Solar Energy 141 (2017) 278–296. https://doi.org/10.1016/j.solener.2016.11.045.

[20] C. Voyant, G. Notton, S. Kalogirou, M.-L. Nivet, C. Paoli, F. Motte, A. Fouilloy, Machine learning methods for solar radiation forecasting: A review, Renewable Energy 105 (2017) 569–582. https://doi.org/10.1016/j.renene.2016.12.095.

[21] X. Luo, D. Zhang, X. Zhu, Deep learning based forecasting of photovoltaic power generation by incorporating domain knowledge, Energy 225 (2021) 120240. https://doi.org/10.1016/j.energy.2021.120240.

[22] E. Fakhraian, M.A. Forment, F.V. Dalmau, A. Nameni, M.J.C. Guerrero, Determination of the urban rooftop photovoltaic potential: A state of the art, Energy Reports 7 (2021) 176–185. https://doi.org/10.1016/j.egyr.2021.06.031.

[23] A.A.A. Gassar, S.H. Cha, Review of geographic information systems-based rooftop solar photovoltaic potential estimation approaches at urban scales, Applied Energy 291 (2021) 116817. https://doi.org/10.1016/j.apenergy.2021.116817.

[24] P. Gagnon, R. Margolis, J. Melius, C. Phillips, R. Elmore, Estimating rooftop solar technical potential across the US using a combination of GIS-based methods, lidar data, and statistical modeling, Environ. Res. Lett. 13 (2018) 024027.

https://doi.org/10.1088/1748-9326/aaa554.

[25] A. Walch, R. Castello, N. Mohajeri, J.-L. Scartezzini, A Fast Machine Learning Model for Large-Scale Estimation of Annual Solar Irradiation on Rooftops, in: Proceedings of the ISES Solar World Congress 2019, International Solar Energy Society, Santiago, Chile, 2019: pp. 1–10. https://doi.org/10.18086/swc.2019.45.12.

[26] A. Walch, R. Castello, N. Mohajeri, J.-L. Scartezzini, Big data mining for the estimation of hourly rooftop photovoltaic potential and its uncertainty, Applied Energy 262 (2020) 114404. https://doi.org/10.1016/j.apenergy.2019.114404.

[27] E. Hadi, A. Heidari, Development of an integrated tool based on life cycle assessment, Levelized energy, and life cycle cost analysis to choose sustainable Facade Integrated Photovoltaic Systems, Journal of Cleaner Production 293 (2021) 126117. https://doi.org/10.1016/j.jclepro.2021.126117.

[28] A. Vulkan, I. Kloog, M. Dorman, E. Erell, Modeling the potential for PV installation in residential buildings in dense urban areas, Energy and Buildings 169 (2018) 97–109. https://doi.org/10.1016/j.enbuild.2018.03.052.

[29] Z. Guo, J. Lu, Q. Chen, Z. Liu, C. Song, H. Tan, H. Zhang, J. Yan, TransPV: Refining photovoltaic panel detection accuracy through a vision transformer-based deep learning model, Applied Energy 355 (2024) 122282. https://doi.org/10.1016/j.apenergy.2023.122282.

[30] Z. Guo, Z. Zhuang, H. Tan, Z. Liu, P. Li, Z. Lin, W.-L. Shang, H. Zhang, J. Yan, Accurate and generalizable photovoltaic panel segmentation using deep learning for imbalanced datasets, Renewable Energy 219 (2023) 119471. https://doi.org/10.1016/j.renene.2023.119471.

[31] U. Eicker, R. Nouvel, E. Duminil, V. Coors, Assessing Passive and Active Solar Energy Resources in Cities Using 3D City Models, Energy Procedia 57 (2014) 896–905. https://doi.org/10.1016/j.egypro.2014.10.299.

[32] J. Liu, Q. Wu, Z. Lin, H. Shi, S. Wen, Q. Wu, J. Zhang, C. Peng, A novel approach for assessing rooftop-and-facade solar photovoltaic potential in rural areas using three-dimensional (3D) building models constructed with GIS, Energy 282 (2023) 128920. https://doi.org/10.1016/j.energy.2023.128920.

[33] R. Pueblas, P. Kuckertz, J.M. Weinand, L. Kotzur, D. Stolten, ETHOS.PASSION: An open-source workflow for rooftop photovoltaic potential assessments from satellite imagery, Solar Energy 265 (2023) 112094. https://doi.org/10.1016/j.solener.2023.112094.

[34] N.P. Weerasinghe, R.J. Yang, C. Wang, Learning from success: A machine learning approach to guiding solar building envelope applications in non-domestic market, Journal of Cleaner Production 374 (2022) 133997.

https://doi.org/10.1016/j.jclepro.2022.133997.

[35] C. Vassiliades, R. Agathokleous, G. Barone, C. Forzano, G.F. Giuzio, A. Palombo, A. Buonomano, S. Kalogirou, Building integration of active solar energy systems: A review of geometrical and architectural characteristics, Renewable and Sustainable Energy Reviews 164 (2022) 112482. https://doi.org/10.1016/j.rser.2022.112482.

[36] B. Tang, J. Zhang, L. Yan, Q. Yu, L. Sheng, D. Xu, Data-Free Generalized Zero-Shot Learning, AAAI 38 (2024) 5108–5117. https://doi.org/10.1609/aaai.v38i6.28316.

[37] M. Kutbi, K.-C. Peng, Z. Wu, Zero-shot Deep Domain Adaptation with Common Representation Learning, IEEE Trans. Pattern Anal. Mach. Intell. (2021) 1–1. https://doi.org/10.1109/TPAMI.2021.3061204.

[38] A. Radford, J.W. Kim, C. Hallacy, A. Ramesh, G. Goh, S. Agarwal, G. Sastry, A. Askell, P. Mishkin, J. Clark, G. Krueger, I. Sutskever, Learning Transferable Visual Models From Natural Language Supervision, in: Proceedings of the 38th International Conference on Machine Learning, PMLR, 2021: pp. 8748–8763. https://proceedings.mlr.press/v139/radford21a.html (accessed May 15, 2024).

[39] A. Kirillov, E. Mintun, N. Ravi, H. Mao, C. Rolland, L. Gustafson, T. Xiao, S. Whitehead, A.C. Berg, W.-Y. Lo, P. Dollár, R. Girshick, Segment Anything, (2023). http://arxiv.org/abs/2304.02643 (accessed April 25, 2024).

[40] J. Ma, Y. He, F. Li, L. Han, C. You, B. Wang, Segment anything in medical images, Nat Commun 15 (2024) 654. https://doi.org/10.1038/s41467-024-44824-z.

[41] Y. Li, D. Wang, C. Yuan, H. Li, J. Hu, Enhancing Agricultural Image Segmentation with an Agricultural Segment Anything Model Adapter, Sensors 23 (2023) 7884. https://doi.org/10.3390/s23187884.

[42] W. Jinlei, C. Chen, C. Dai, J. Hong, A Domain-Adaptive segmentation method based on segment Anything model for mechanical assembly, Measurement (2024) 114901. https://doi.org/10.1016/j.measurement.2024.114901.

[43] F.A. Pramadya, K.N. Kim, Promoting residential rooftop solar photovoltaics in Indonesia: Net-metering or installation incentives?, Renewable Energy 222 (2024) 119901. https://doi.org/10.1016/j.renene.2023.119901.

[44] A.C. Lemay, S. Wagner, B.P. Rand, Current status and future potential of rooftop solar adoption in the United States, Energy Policy 177 (2023) 113571. https://doi.org/10.1016/j.enpol.2023.113571.

[45] T. Gómez-Navarro, T. Brazzini, D. Alfonso-Solar, C. Vargas-Salgado, Analysis of the potential for PV rooftop prosumer production: Technical, economic and

environmental assessment for the city of Valencia (Spain), Renewable Energy 174 (2021) 372–381. https://doi.org/10.1016/j.renene.2021.04.049.

[46] B. Guan, H. Yang, T. Zhang, X. Liu, X. Wang, Technoeconomic analysis of rooftop PV system in elevated metro station for cost-effective operation and clean electrification, Renewable Energy 226 (2024) 120305. https://doi.org/10.1016/j.renene.2024.120305.

[47] Y. Fu, W. Xu, Z. Wang, S. Zhang, X. Chen, X. Zhang, Experimental study on thermoelectric effect pattern analysis and novel thermoelectric coupling model of BIPV facade system, Renewable Energy 217 (2023) 119055. https://doi.org/10.1016/j.renene.2023.119055.

[48] X. Li, K. Li, Y. Sun, R. Wilson, J. Peng, K. Shanks, T. Mallick, Y. Wu, Comprehensive investigation of a building integrated crossed compound parabolic concentrator photovoltaic window system: Thermal, optical and electrical performance, Renewable Energy 223 (2024) 119791. https://doi.org/10.1016/j.renene.2023.119791.

[49] X. Su, L. Zhang, Y. Luo, Z. Liu, Energy performance of a reversible window integrated with photovoltaic blinds in Harbin, Building and Environment 213 (2022) 108861. https://doi.org/10.1016/j.buildenv.2022.108861.

[50] W. F. Holmgren, C. W. Hansen, M. A. Mikofski, pvlib python: a python package for modeling solar energy systems, JOSS 3 (2018) 884. https://doi.org/10.21105/joss.00884.

[51] K. Branker, M.J.M. Pathak, J.M. Pearce, A review of solar photovoltaic levelized cost of electricity, Renewable and Sustainable Energy Reviews 15 (2011) 4470–4482. https://doi.org/10.1016/j.rser.2011.07.104.

[52] C.S. Lai, M.D. McCulloch, Levelized cost of electricity for solar photovoltaic and electrical energy storage, Applied Energy 190 (2017) 191–203. https://doi.org/10.1016/j.apenergy.2016.12.153.

[53] H. Sharadga, S. Hajimirza, R.S. Balog, Time series forecasting of solar power generation for large-scale photovoltaic plants, Renewable Energy 150 (2020) 797–807. https://doi.org/10.1016/j.renene.2019.12.131.

[54] M. Jiang, J. Li, W. Wei, J. Miao, P. Zhang, H. Qian, J. Liu, J. Yan, Using Existing Infrastructure to Realize Low-Cost and Flexible Photovoltaic Power Generation in Areas with High-Power Demand in China, iScience 23 (2020) 101867. https://doi.org/10.1016/j.isci.2020.101867.

[55] M. Wang, X. Mao, Y. Gao, F. He, Potential of carbon emission reduction and financial feasibility of urban rooftop photovoltaic power generation in Beijing, Journal of Cleaner Production 203 (2018) 1119–1131.

https://doi.org/10.1016/j.jclepro.2018.08.350.